\documentclass[10pt,twocolumn,letterpaper]{article}

\usepackage{iccv}              

\definecolor{iccvblue}{rgb}{0.21,0.49,0.74}
\usepackage[pagebackref,breaklinks,colorlinks,allcolors=iccvblue]{hyperref}

\usepackage{multirow}

\def\paperID{2320} 
\def\confName{ICCV}
\def\confYear{2025}

\title{Object-Aware Video Matting with Cross-Frame Guidance}

\author{
    Huayu Zhang\textsuperscript{\rm 1}, Dongyue Wu\textsuperscript{\rm 1}, Yuanjie Shao\textsuperscript{\rm 1}, Nong Sang\textsuperscript{\rm 1}, Changxin Gao\textsuperscript{\rm 1}\thanks{Corresponding author.}\\
    \textsuperscript{\rm 1}National Key Laboratory of Multispectral Information Intelligent Processing Technology, \\
    School of Artificial Intelligence and Automation, Huazhong University of Science and Technology
}

\begin{document}
\maketitle
\begin{abstract}
Recently, trimap-free methods have drawn increasing attention in human video matting due to their promising performance. Nevertheless, these methods still suffer from the lack of deterministic foreground-background cues, which impairs their ability to consistently identify and locate foreground targets over time and mine fine-grained details. In this paper, we present a trimap-free Object-Aware Video Matting (OAVM) framework, which can perceive different objects, enabling joint recognition of foreground objects and refinement of edge details. Specifically, we propose an Object-Guided Correction and Refinement (OGCR) module, which employs cross-frame guidance to aggregate object-level instance information into pixel-level detail features, thereby promoting their synergy. Furthermore, we design a Sequential Foreground Merging augmentation strategy to diversify sequential scenarios and enhance capacity of the network for object discrimination. Extensive experiments on recent widely used synthetic and real-world benchmarks demonstrate the state-of-the-art performance of our OAVM with only an initial coarse mask. The code and model will be available.
\end{abstract}    
\section{Introduction}
\label{sec:intro}

Video matting aims at predicting alpha mattes of each frame from an input video, which can be widely used in applications like virtual reality and film production. Each frame ${I_k}$ in a natural video sequence $\mathbf{I}= \left\{ {{I_1},{I_2},...,{I_T}} \right\}$ can be represented as a linear combination of foreground ${F_k}$ and background ${B_k}$ with alpha matte ${\alpha_k}$ by a matting equation:
\begin{equation}
{I_k} = {\alpha _k}{F_k} + (1 - {\alpha _k}){B_k},{\alpha _k} \in [0,1].
\end{equation}
Matting is a highly ill-posed problem, due to the need to solve 7 unknown variables per pixel from only 3 known values. A classic solution is to use the trimap as an auxiliary input \cite{DVM,TCVOM}. The trimap explicitly labels the foreground, background, and unknown regions of the image, allowing the matting network to focus on fewer unknown regions. While this solution has yielded favourable results, labelling trimap is labour-intensive and time-consuming, which is difficult for novices to practice.

It is noticeable that there are also some trials \cite{modnet,rvm,vmformer} to get rid of trimaps to predict alpha mattes. But the absence of explicit location cues means that they may confuse foreground objects with the background, making video matting performance unsustainable. Using an initial coarse mask as the auxiliary input can somewhat alleviate this problem \cite{adam,HSTSG}. The mask is provided only for the first frame of the video for accurately predicting the alpha mattes for all subsequent frames. However, the inaccuracy of the given coarse mask makes the model tend to ignore it and determine the foreground and background by itself during relatively complex scenes. This often leads to predicted foreground range variations, making it difficult to lock on to the specific objects and refine them in video clips.

\begin{figure}[t]
\centering
\includegraphics[width=0.98\columnwidth]{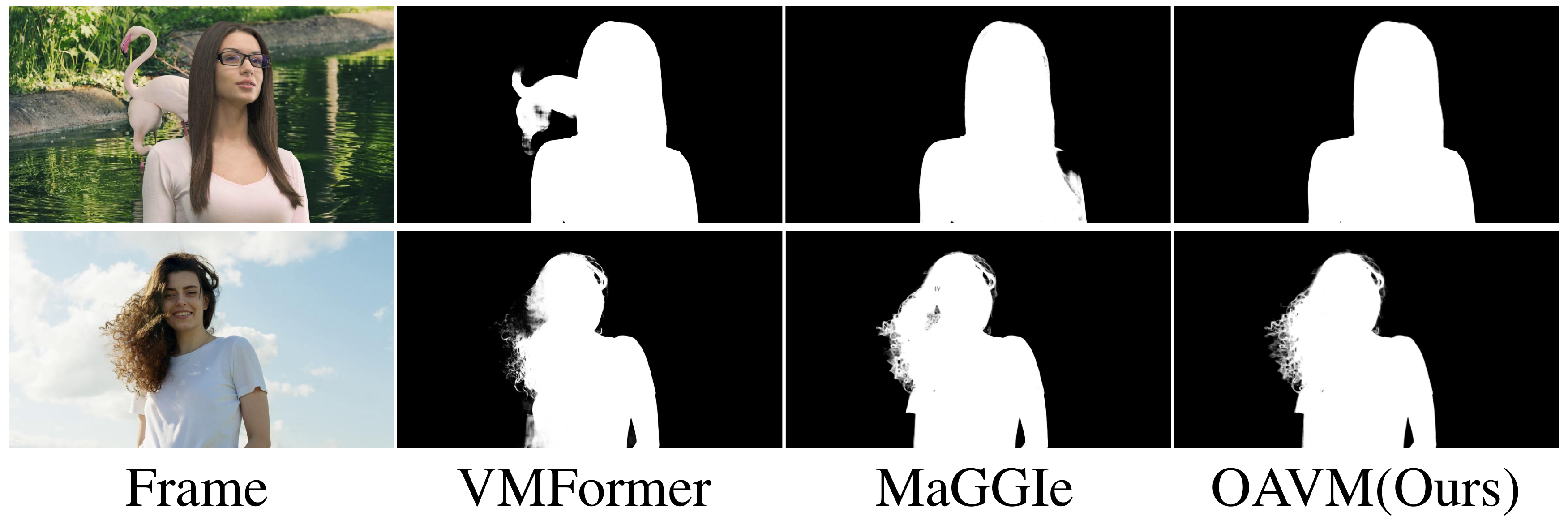}
\caption{We explore endowing the network with the ability to jointly recognize foreground targets and refine edge details without trimaps. As illustrated in the example, our method consistently delivers outstanding performance across challenging video cases, whether it is with similar foreground and background colors or very detailed hair.}
\label{first_fig}
\end{figure}

In this paper, we explore the possibility of endowing the network with the ability to jointly recognize foreground targets and refine edge details without trimaps, and present an Object-Aware Video Matting (OAVM) framework that can perceive objects for matting when processing video frames. Specifically, we first extract pixel-level temporal features based on cross-frame matching and object-level instance queries based on set prediction, respectively. We argue that capturing object-level information is crucial for video matting, which can assist to distinguish foreground objects from background. And then, we introduce an Object Guided Correction and Refinement (OGCR) module to harmoniously integrate the two levels of information, so that pixel features are rectified with object-level queries to achieve synergy. Due to the lack of understanding of foreground and background in pixel-level information, and the lack of fine-grained details in object-level information, we propose cross-frame guidance to induce pixel feature perception of objects in a sequence with instance queries, thereby focusing on foreground targets. Subsequently, the foreground objects are further refined in the module to boost the quality of details. Furthermore, to improve the composition procedure of  video clips during training, we design a Sequential Foreground Merging data augmentation strategy, which can make the scenarios more diverse and enhance the object discrimination ability of network from the training data level. As shown in the example of Figure \ref{first_fig}, in challenging video cases, whether with similar foreground and background colors or very detailed hair, our method is capable of accurately identifying the foreground and refining the detailed parts.

We conduct extensive experiments on widely used synthetic datasets (RVM Benchmark \cite{rvm} and VMFormer Benchmark \cite{vmformer}) and real-world datasets (CRGNN Benchmark \cite{crgnn}) for video matting. The results show that our method outperforms previous methods and achieves the state-of-the-art level. Specifically, our method attains 4.23, 0.31 and 3.87 in MAD, MSE, and Grad in RVM Benchmark without manual efforts. Main contributions of this paper are summarized as below.

\begin{itemize}
\item For the first time, we explore empowering video matting networks with the ability to understand object-level information explicitly and present the Object-Aware Video Matting (OAVM) framework, which can jointly identify foreground objects and refine edge details for trimap-free video matting.
\item We propose the Object-Guided Correction and Refinement (OGCR) module to induce pixel features to perceive objects. We design cross-frame guidance to aggregate object-level instance information into pixel-level detail features, thereby promoting their synergy.
\item We streamline the training procedure for video matting and design the Sequential Foreground Merging augmentation strategy to improve the composition procedure of video clips, which enhances the object discrimination ability of the network from the training data level.
\item Extensive experiments on recent widely used synthetic and real-world benchmarks demonstrate the superior performance of our framework.
\end{itemize}
\section{Related Work}
\label{sec:formatting}

After the development of traditional methods \cite{bayesian,Poisson,closedform,sharedsampling}, a number of deep learning-based matting methods have been proposed with promising results, and they can be divided into trimap-based methods and trimap-free methods. Trimap-based matting methods \cite{DIM,indexmat,gca,Matteformer,Transmatting,ViTMatte} employ the user-defined trimaps, which providing clear guidance for the network. For the trimap-free methods, researchers explored finding alternatives \cite{mg,mgWild} or using no auxiliary input \cite{latefusion,Boosting,hattmatting,GFM}. Recently, there are also some researches focusing on leveraging generic image segmentation models to assist the matting process \cite{Matteanything,MAM}.



\paragraph{Video Matting.}

Besides focusing on fine-grained detail extraction, video matting also emphasizes maintaining spatial and temporal consistency versus image matting. Deep video matting can also be divided into trimap-based and trimap-free methods. Trimap-based methods typically propagate the trimap between frames \cite{DVM,TCVOM,OTVM,FTPVM}. The annotation of trimap consumes significant time and labor expenses, hence trimap-free methods have also received a great deal of attention \cite{modnet,bgm,bgmv2240k,rvm,vmformer,BiMatting}. AdaM \cite{adam} uses the mask of the first frame as an initial condition and takes advantage of both spatial and motion cues. MaGGIe \cite{maggie} uses the mask of each frame as prior and enhances temporal consistency at the feature and matte level. Nonetheless, trimap-free video matting methods still struggle to identify foregrounds in the sequence. In our work, object-level instance information is integrated into the video matting procedure. With the proposed cross-frame guidance, the object-level features are transformed into foreground-background prompting, which aids in localizing the target object and avoiding foreground misjudgments.


\paragraph{Set Prediction.}

\begin{figure*}[t]
\centering
\includegraphics[width=0.96\textwidth]{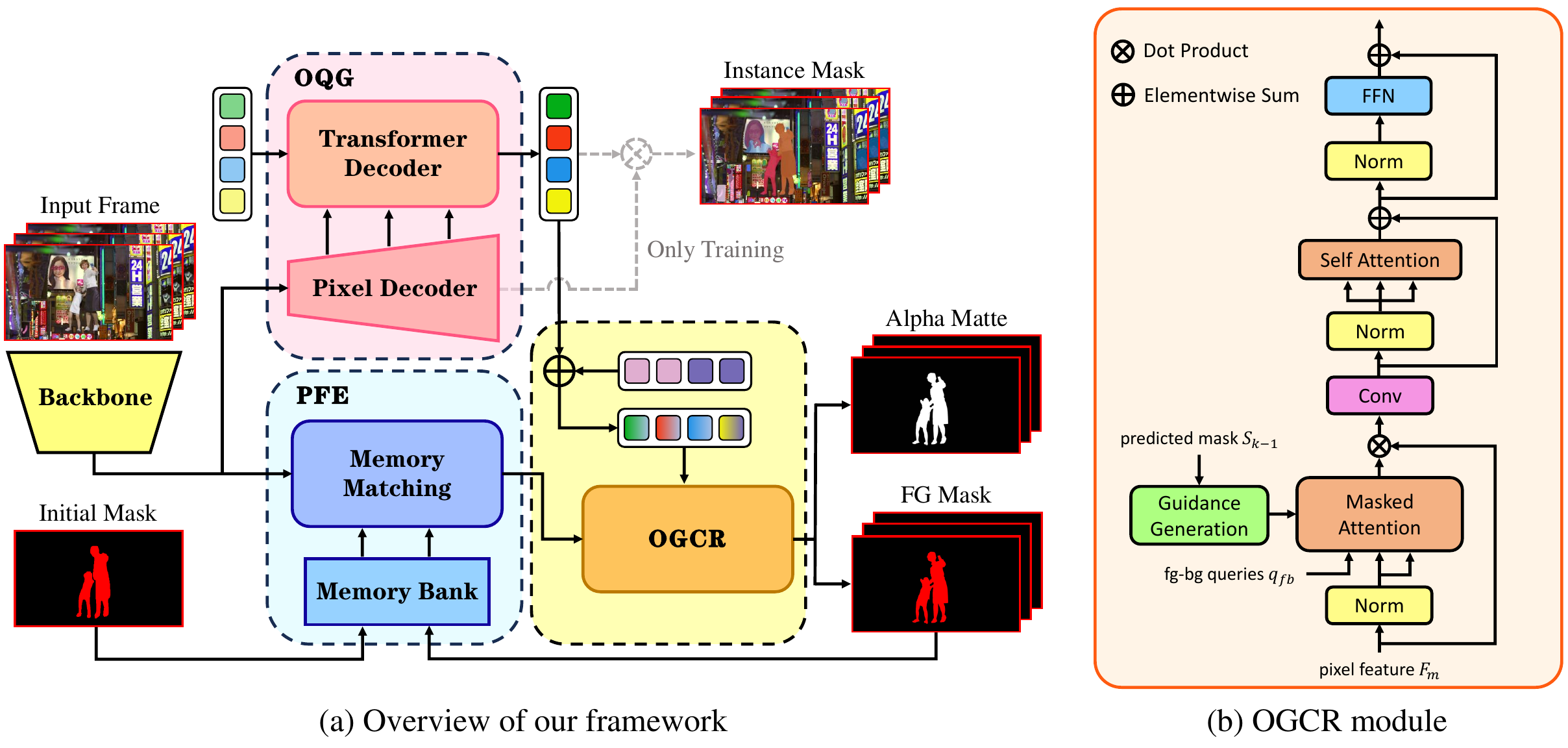}
\caption{(a) The overview of the proposed OAVM framework. Pixel-level Feature Extraction (PFE) and Object-level Query Generation (OQG) are implemented to yield features with temporal coherence and queries embedding object-level instances, respectively. Subsequently, the two are integrated by the Object-Guided Correction and Refinement (OGCR) module, where the detailed features are further refined, ultimately producing the alpha matte and foreground mask of the current frame. The instance mask output is used to optimize the network during the training phase. (b) Illustration of the Object-Guided Correction and Refinement (OGCR) module.}
\label{overview}
\end{figure*}

Set prediction, introduced with DETR \cite{detr}, has been used in a wide range of vision tasks. Mask2Former \cite{m2f} employs this concept for segmentation and achieves good results. Mask DINO \cite{Maskdino} extends DINO \cite{Dino} and unifies detection and segmentation. However, directly applying set prediction to matting does not yield satisfactory performance due to the inherent differences between segmentation \cite{XMem,DEAOT,ISVOS} and matting \cite{FTPVM,adam} tasks. While segmentation focuses on distinguishing objects with different semantics, matting is concerned with extracting the fine details of the foreground. Our method introduces set prediction into matting through a novel design, endowing the network with object-awareness. This design enables the network to locate the foreground and facilitate fine-grained feature extraction, even in the absence of explicit target guidance (i.e., trimaps).

\section{Methodology}

We aim to endow the video matting network with object-awareness, enabling it to precisely orient the foreground target and focus more on fine-grained features for better semantic localization and detail extraction. The architecture of the Object-Aware Video Matting (OAVM) is illustrated in Figure \ref{overview}. Simply given an initial coarse mask predicted by any off-the-shelf segmentation network, we process the frames sequentially and output the foreground mask and alpha matte for each frame. The memory bank is employed to store past frames and their predicted masks.

Formally, for the current frame ${I_k} \in \mathbf{I}$ in the video sequence, we first extract a pyramid of feature maps $\mathbf{F_k} = \{ {F_{s1}},{F_{s2}},{F_{s3}},{F_{s4}}\}$ from the backbone. The feature ${F_{s4}}$ are performed in parallel for pixel-level feature extraction based on memory matching and object-level query generation based on set prediction to produce the features ${F_m}$ with temporal coherence and queries ${q_o}$ embedding object-level instances, respectively. Next, the Object Guided Correction and Refinement (OGCR) module enables to harmoniously integrate ${F_m}$ and ${q_o}$ so that pixel features are rectified with object-level information. Finally, the corrected and refined features ${F_r}$ and multi-scale features from backbone are fed into the matting decoder to produce the foreground mask ${S_k}$ and alpha matte ${\alpha _k}$ for the current frame.

\subsection{Pixel-level Temporal Feature Extraction}

It is a central challenge in video matting to ensure temporal coherence and prevent the degradation of matting quality over time. We construct a memory bank to store the initial and historical foreground embeddings and regress each pixel of the current frame according to their similarities to the target foreground in the memory bank. 

The attention mechanisms perform pixel-level matching between the current frame and the memorized frames. Specifically, $Q \in {\mathbb{R}^{H \times W \times C}}$, $K \in {\mathbb{R}^{T \times H \times W \times C}}$, $V \in {\mathbb{R}^{T \times H \times W \times C}}$ are defined as the query embedding of the current frame, the key embedding, and the value embedding of the memorized frames respectively, which are computed by linear projection. Where $T$, $C$ denote the temporal and channel dimensions, and $H$, $W$ denote the height and width dimensions after downsampling. Subsequently, frame-to-frame attention operations can be expressed as:
\begin{equation}
FFAttn(Q,K,V,M) = {\rm{softmax}}(\frac{{Q{K^T}}}{{\sqrt C }})(V + FE(S)),
\end{equation}
where $S \in {\mathbb{R}^{T \times H \times W \times 1}}$ is the predicted foreground masks of the memory frames. $FE$ is a foreground embedding layer. In our approach, two forms of attention are employed to model the global and local relationships between frames for the foreground matching and propagation. We treat earlier frames as long-term memory and more recent frames as short-term memory. For long-term memory, the global attention is designed due to the wide range of foreground variations:

\begin{equation}
{F_G} = GFFAttn(Q,K,V) = FFAttn(Q,K,V,M).
\end{equation}

For short-term memory, the local attention can capture matching relations more efficiently and accurately based on small shifts of the foreground between adjacent frames. The attention is restricted to a fixed window size. For a pixel $i$ in the current frame, the region it attends to is confined within $W(i)$, where $W(i)$ is a spatial neighborhood of size $w \times w$ centered on $i$. The local attention is computed as:

\begin{equation}
\begin{aligned}
{F_L} &= LFFAttn(Q,K,V) \\
&= FFAttn({Q_i},{K_{W(i)}},{V_{W(i)}},{M_{W(i)}}).
\end{aligned}
\end{equation}
${F_G}$ and ${F_L}$ are then combined to produce the final matched output ${F_m}$. In practice, we use only the first frame for long-term attention and the last frame for short-term attention mechanisms. This minimizes the memory storage occupation and lowers the cost. When performing matting on the first frame, there is no additional memory information available for matching, hence self-attention is performed instead.

\subsection{Object-level Query Generation}

Previous video matting networks suffer from weak discrimination of objects, which potentially result in misjudging the background regions in the video and also interfere with the capture of fine-grained details in foreground regions. We investigate how to mine object-level information and use it for correcting and refining pixel-level matching features. Inspired by \cite{detr,m2f}, we build an object-level query generation module that is capable of performing set prediction for instance segmentation.

Specifically, we construct a lightweight pixel decoder, which is an iterative upsampling architecture that takes $\mathbf{F}$ as input and progressively propagates low-level features through convolutional and upsampling layers alternately to obtain multi-scale features $\{ {F_{Pi}}\} _{i = 0}^2$. Next, a transformer decoder is designed which encodes the image feature information into a set of learnable object queries $q \in {\mathbb{R}^{N \times C}}$, where $N$ is the predefined number of queries. This set of object queries is learnable, interacting with multi-scale features layer by layer through masked attention.

Subsequently, self-attention and multi-layer perception are applied to output queries ${q_o}$ that imply object-level comprehension. The total number of iterative transformer decoder layers $L$ is set to 3, i.e., the query interacts with each scale feature only once.

\subsection{Object Guided Correction and Refinement Module}

After pixel-level temporal features and object-level instance queries are extracted, how to make the two synergize in a reasonable manner is worth exploring. We perform mutual information exchange between them with the OGCR module and propose cross-frame guidance in order to incorporate the object queries, which contain valid foregrounds, into the matting procedure, thus enabling the network to accurately localize the foreground target and be more concerned with the fine-grained details. The illustration of the OGCR module is shown in Figure \ref{overview}. 

In general, transitions between scenes in video are temporally smooth, and the same object will appear at similar local locations between adjacent frames. Based on this property, we generate the cross-frame guidance as foreground localization prompts for the current frame using the predicted mask of the last frame ${S_{k-1}}$. Leveraging queries ${q_o}$ with the cross-frame guidance, we further perform the correction and refinement to ${F_m}$.

Given the domain discrepancy between object features and foreground features, the transformation between them is essential. We design a learnable embedding ${E_{fb}}$ to mine foreground-background knowledge. When combined with object-level queries, this embedding effectively translates instance information into the foreground-background information required for accurate matting. The object queries is fused to the learnable embedding ${E_{fb}}$ by a linear addition operation:
\begin{equation}
{q_{fb}} = {E_{fb}} + {q_o}({I_k}).
\end{equation}
The foreground-background queries ${q_{fb}}$ is then updated by attending to the pixel feature ${F_m}$. The normal global attention would leave queries without any foreground hints attached, resulting in all object features being sent back equally to the ${F_m}$. Therefore, we propose the cross-frame guidance when updating queries, which allows queries to understand the concepts of foreground and background, thus adapting to the matting task requirements. In our frame-to-query cross-attention, masked attention with cross-frame guidance is presented, which computes as:
\begin{equation}
\begin{aligned}
{X_m} &= FQAttn({Q_q},{K_f},{V_f},{M_S}) \\
&= {\rm{softmax}} (\frac{{{Q_q}K_f^T}}{{\sqrt C }} + {M_S}){V_f},
\end{aligned}
\end{equation}

where ${Q_q}$ is a linear projection of ${q_{fb}}$ and ${K_f}$, ${V_f}$ are linear projections of ${F_m}$. The key feature was also added with a fixed 2D sinusoidal positional embedding for each resolution. ${M_S}$ is obtained via a cross-frame guidance generation operation. After resampling the mask of the previous frame, ${M_S}$ is generated by dilation $\mathcal{D}$ and binarization $\mathcal{B}$ with the threshold of 0.5, which at feature location $(x,y)$ is:
\begin{equation}
{M_{S,k}}(x,y) = \left\{ {\begin{array}{*{20}{l}}
{0,}&{{\rm{if \; \mathcal{D}\& \mathcal{B}(}}{S_{k - 1}}{\rm{)}}(x,y){\rm{ = 1}}}\\
{ - \infty ,}&{{\rm{otherwise}}}
\end{array}} \right.,
\end{equation}

where $k$ is the index of frames in the sequence. With this masked attention guided by cross-frame, we facilitate the separation of semantics between foreground and background in preparation for object-level correction of temporal pixel features. Next, the ${X_m}$ is called back to the pixel features to aggregate the foreground object information into the pixel features and correct the foreground misjudgment. The dot product between ${X_m}$ and ${F_m}$ is performed, and the output is concatenated with ${F_m}$ to finally produce the object-aware pixel features.

The cross-frame guidance only provides the approximate position of the foreground object and can't precisely capture the edge detail information that is required for matting. For pixel-level edge reasoning, we further perform self-attention and feed-forward layers. After the Object Guided Correction and Refinement, object-level instance queries and pixel-level detail features are rationally integrated and further enhanced to output object-aware fine-grained features ${F_o}$.

\subsection{Prediction Stage}

Instance segmentation supervision is also applied in order to explicitly acquire object comprehension in addition to matting supervision. To guide the network for object-level comprehension, we build the prediction stage using a mask classification strategy similar to \cite{mf}, where binary masks are decoded from per-pixel embedding and object queries after the pixel decoder and transformer decoder, as a prediction of instance objects. An auxiliary loss is employed in every intermediate transformer decoder layer and in the learnable query features before the transformer decoder. Note that the set prediction part will only be used during training and discarded during inference. To predict the final alpha mattes and  foreground masks, the matting decoder is built after the OGCR module, which mainly comprises layers of convolution, Batch Normalization and activation to recover the detail information in the pixel features ${F_o}$ and the feature pyramid $\mathbf{F_k}$.

\subsection{Sequential Foreground Merging Augmentation}

Previous video matting methods have struggled to consistently achieve good results in multi-object sequence scenarios. We argue that one of the reasons is the overfitting to training data that contains single portraits. This leads the model to focus on separate foregrounds during training, which is not conducive to learning object relationships that often occur in the real world, such as occlusion and parallelism. To improve the data pipeline for generating training samples, we designed the Sequential Foreground Merging augmentation for video matting. Our strategy enables more diverse background information and enhances the object discrimination ability of the network from the data level.

Specifically, for a foreground sequence $\{ {F_1}\} $ and a background sequence $\{ B\} $, we randomly pick another foreground sequence $\{ {F_2}\} $ from the training set with probability ${p_1}$, combine it with $\{ B\} $ in the form of frame-by-frame correspondence to compose a new background sequence $\{ {B_n}\} $, which is subsequently synthesized with $\{ {F_1}\} $ to obtain the desired training video clip:
\begin{equation}
\{ {B_n}\}  = \{ {\alpha _2}\}  \odot \{ {F_2}\}  + (1 - \{ {\alpha _2}\} ) \odot \{ B\} ,
\end{equation}
\begin{equation}
\{ {I_n}\}  = \{ {\alpha _1}\}  \odot \{ {F_1}\}  + (1 - \{ {\alpha _1}\} ) \odot \{ {B_n}\} ,
\end{equation}
where $ \odot $ denotes element-wise multiplication, $\{ {\alpha _1}\}$ and $\{ {\alpha _2}\}$ are the alpha matte sequences corresponding to $\{ {F_1}\} $ and $\{ {F_2}\} $. Moreover, the $\{ {\alpha _i}\}$ are binarized with a threshold of 0.5 to obtain $\{ {M_i}\} $, which serve as the GT masks for the foregrounds. After modulation, the supervision $\{ {\alpha _{GT}}\}$ for video matting is randomly selected by:
\begin{equation}
\{ {\alpha _{GT}}\} = \left\{ {\begin{array}{*{20}{l}}
{\{ {\alpha _1}\} ,}&{{\rm with} \: {p_2}}\\
{1 - (1 - \{ {\alpha _1}\} ) \odot (1 - \{ {\alpha _2}\} ),}&{{\rm with} \: 1-{p_2}}
\end{array}} \right.,
\end{equation}
and $\{ {M_i}\} $ are used as the supervision for instance segmentation. Our augmentation strategy enriches the diversity of scenarios and enables the model to improve object discrimination capabilities.
\section{Experiments}

\subsection{Datasets and Implementation Details}

\begin{table*}[t]
\centering
\scalebox{1.0}{
\begin{tabular}{*{11}{c}}
  \toprule[0.5mm]
  \multirow{2}{*}{Method} & \multirow{2}{*}{Backbone} & \multicolumn{5}{c}{RVM 512×288} & \multicolumn{4}{c}{RVM 1920×1080} \\
  \cmidrule(lr){3-7}\cmidrule(lr){8-11}
  && MAD & MSE & Grad & Conn & dtSSD & MAD & MSE & Grad & dtSSD \\
  \midrule
  \multicolumn{11}{l}{\textit{Auxiliary-free Methods}} \\
  \midrule
  MODNet\cite{modnet} & MobileNet & 9.41 & 4.30 & 1.89 & 0.81 & 2.23 & 11.13 & 5.54 & 15.3 & 3.08 \\
  RVM\cite{rvm} & MobileNet & 6.08 & 1.47 & 0.88 & 0.41 & 1.36 & 6.57 & 1.93 & 10.55 & 1.90 \\
  VMFormer\cite{vmformer} & MobileNet & 6.02 & 1.00 & 0.75 & 0.37 & - & 6.20 & 1.53 & 6.30 &  - \\
  \midrule
  \multicolumn{11}{l}{\textit{Auxiliary-based Methods}} \\
  \midrule
  BGMv2\cite{bgmv2240k} & MobileNet & 25.19 & 19.63 & 2.28 & 3.26 & 2.74 & - & - & - & - \\
  FBA\cite{fbmatting} & ResNet & 8.36 & 3.37 & 2.09 & 0.75 & 2.09 & - & - & - & - \\
  FTP-VM\cite{FTPVM} & ResNet & 6.09 & 1.28 & 1.13 & 0.39 & 1.57 & 7.83 & 3.16 & 21.37 & 2.11 \\
  AdaM\cite{adam} & MobileNet & 5.30 & 0.78 & 0.72 & 0.30 & 1.33 & 4.42 & 0.39 & 5.12 & 1.39 \\
  MaGGIe\cite{maggie} & ResNet & 5.57 & 0.68 & 0.63 & 0.35 & 1.34 & 4.55 & 0.46 & 4.67 & 1.38 \\
  OAVM (Ours) & MobileNet & \textbf{4.92} & \textbf{0.62} & \textbf{0.57} & \textbf{0.23} & \textbf{1.22} & \textbf{4.23} & \textbf{0.31} & \textbf{3.87} & \textbf{1.31} \\
  \bottomrule[0.5mm]
\end{tabular}}
\caption{\label{rvmcompare}The quantitative evaluation on RVM benchmark. The upper group represents video matting models that do not use auxiliary inputs, while the lower models utilize auxiliary information, such as masks or trimaps. As illustrated, our OAVM shows the state-of-the-art performance in both resolutions using only minimal auxiliary information (the coarse mask for the first frame).}
\end{table*}

\begin{table*}[t]
\centering
\scalebox{1.0}{
\begin{tabular}{*{11}{c}}
  \toprule[0.5mm]
  \multirow{2}{*}{Method} & \multirow{2}{*}{Backbone} & \multicolumn{5}{c}{VMF 512×288} & \multicolumn{4}{c}{VMF 1920×1080} \\
  \cmidrule(lr){3-7}\cmidrule(lr){8-11}
  && MAD & MSE & Grad & Conn & dtSSD & MAD & MSE & Grad & dtSSD \\
  \midrule
  \multicolumn{11}{l}{\textit{Auxiliary-free Methods}} \\
  \midrule
  MODNet\cite{modnet} & MobileNet & 10.39 & 5.65 & 2.02 & 1.04 & - & 11.56 & 6.47 & 15.23 & - \\
  RVM\cite{rvm} & MobileNet & 5.99 & 1.17 & 1.10 & 0.34 & - & 6.45 & 1.63 & 11.59 & - \\
  VMFormer\cite{vmformer} & MobileNet & 4.91 & 0.55 &  0.40 & 0.25 & - & 4.81 & 0.78 & 4.90 & 3.34 \\
  \midrule
  \multicolumn{11}{l}{\textit{Auxiliary-based Methods}} \\
  \midrule
  BGMv2\cite{bgmv2240k} & MobileNet & 6.63 & 1.79 & 1.54 & 0.50 & - & 27.01 & 21.31 & 20.34 & - \\
  FTP-VM\cite{FTPVM} & ResNet & 4.98 & 0.76 & 0.43 & 0.31 & 1.47 & 4.86 & 0.83 & 4.92 & 2.78 \\
  AdaM\cite{adam} & MobileNet & 5.12 & 0.87 & 0.63 & 0.32 & 1.78 & 5.22 & 0.97 & 5.21 & 3.75 \\
  MaGGIe\cite{maggie} & ResNet & 4.97 & 0.59 & 0.42 & 0.28 & 1.32 & 4.77 & 0.51 & 4.36 & 2.35 \\
  OAVM (Ours) & MobileNet & \textbf{4.50} & \textbf{0.46} & \textbf{0.34} & \textbf{0.18} & \textbf{1.01} & \textbf{4.01} & \textbf{0.22} & \textbf{3.39} & \textbf{1.18} \\
  \bottomrule[0.5mm]
\end{tabular}}
\caption{\label{vmformercompare}The quantitative evaluation on VMFormer benchmark. The upper group represents video matting models that do not use auxiliary inputs, while the lower models utilize auxiliary information, such as masks or trimaps. As illustrated, our OAVM shows the state-of-the-art performance in both resolutions using only minimal auxiliary information (the coarse mask for the first frame).}
\end{table*}

\paragraph{Datasets.}

Notably, although we design object-level supervision for video matting pipeline, we do not use any additional datasets in comparison with previous work. We use video matting datasets \cite{bgmv2240k}, video segmentation datasets \cite{youtubevos}, and image segmentation datasets \cite{coco} for training and choose the background from DVM \cite{DVM} and BG20K \cite{GFM}. It is expected that our model will be further enhanced if additional training is performed on the AIM \cite{DIM} and D646 \cite{hattmatting} datasets. We evaluate the performance on the broadly used synthetic (RVM \cite{rvm}, VMFormer \cite{vmformer}) and real-world (CRGNN \cite{crgnn}) benchmarks. 





\paragraph{Network Details.}

The initial coarse mask can be predicted from any off-the-shelf segmentation network. To ensure a fair comparison, we use Mask-RCNN \cite{maskrcnn} with ResNet50 \cite{resnet}, the same as in AdaM \cite{adam}. We choose MobileNet \cite{mobilenetv3} as our backbone to improve video inference speed. The feature dimension in the object-level query generation branch is set to 128. The window size $w$ is set to 15. Moreover, the probabilities $p_1$, $p_2$ in the augmentation are 0.4 and 0.5.


\paragraph{Training Details.}

We streamline the training process and divide it into two stages. In the first stage, we train the model alternatively on VM240K \cite{bgmv2240k}, COCO \cite{coco} and YouTubeVOS \cite{youtubevos} at low-resolution for 15 epochs. The learning rate is set to $2 \times {10^{ - 4}}$ and the weight decay is set to 0.07. In the second stage, The model is trained on VM240K at high-resolution for 2 epochs, with the learning rate of $1 \times {10^{ - 5}}$ and the weight decay of 0.07. The Sequential Foreground Merging augmentation is applied to the video matting dataset.

\subsection{Comparison with the State-of-the-art Methods}

\paragraph{Results on Synthetic Benchmarks.}

\begin{figure*}[t]
\centering
\includegraphics[width=0.98\textwidth]{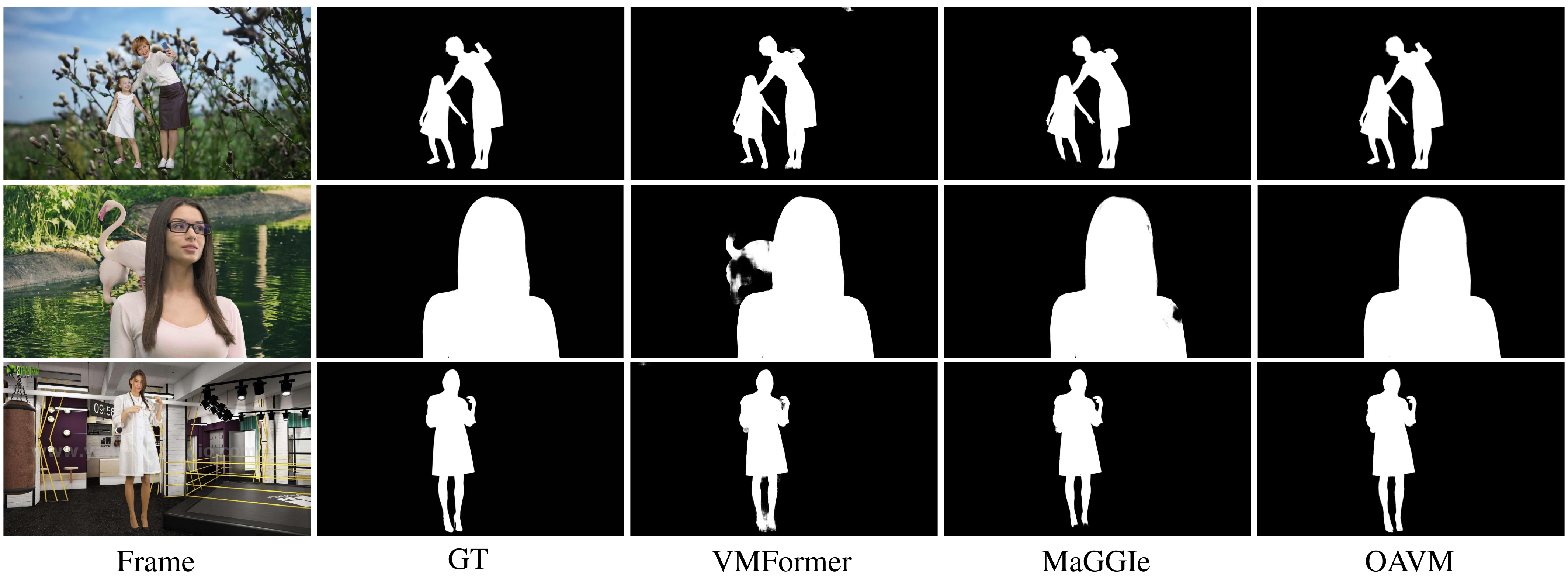}
\caption{Qualitative comparisons on on the RVM benchmark. Our OAVM yields excellent results in both edge detail and center regions. Please zoom in for a better view.}
\label{rvm_visual}
\end{figure*}

The comparison results between OAVM and existing state-of-the-art models on the RVM Benchmark are listed in Table \ref{rvmcompare}. For all metrics, lower values indicate better performance. Our model faithfully utilizes the initial coarse mask and performs trimap-free video matting. We observe that even though our method uses only minimal auxiliary information and adopts a more simplified training pipeline, OAVM surpasses all the competitors in both resolutions, reaching 4.23, 0.31 and 1.31 on MAD, MSE and dtSSD. Experiments are also conducted on the VMFormer benchmark and results are summarized in Table \ref{vmformercompare}. Our method achieves superior performance with a clear margin over the others. We argue that this is attributed to our network's capability to understand both object-level and pixel-level information, correcting and guiding detail extraction through object-level queries. Moreover, some qualitative results are shown in Figure \ref{rvm_visual}. In some complex scenes (e.g., sequences with people interaction), our method precisely localize the target and extract detail features, showing outstanding visual performance. 



\begin{figure}[t]
\centering
\includegraphics[width=0.98\columnwidth]{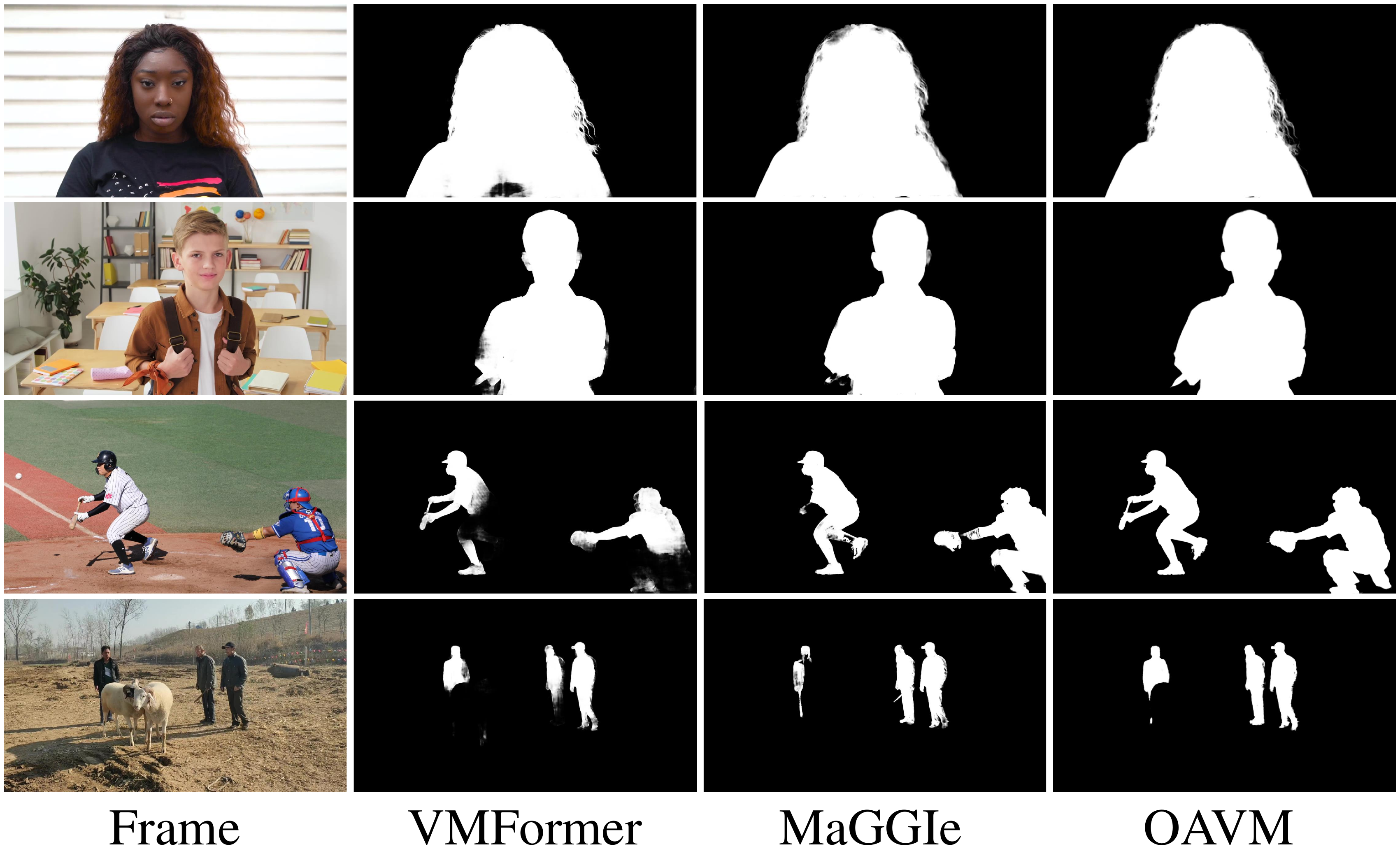}
\caption{Visual comparisons on the real-wold scenarios. Frames in the first two rows originate from the CRGNN benchmark, and those in the subsequent two rows are sourced from DAVIS and MOSE datasets. Please zoom in for a better view.}
\label{real_visual}
\end{figure}

\paragraph{Results on Real-world Benchmarks.}

We validate the performance on real-world datasets without additional fine-tuning for evaluating generalization capability. The quantitative results are shown in Table \ref{crgnncompare} and some visualizations are provided in Figure \ref{real_visual}. Our method performs better in discriminating between different objects and clearly extracting foreground boundaries, effectively overcoming the domain shift problem and generalizing well to real-world scenarios. 


\paragraph{More Comparative Studies.}

\begin{table}[t]
\setlength{\tabcolsep}{7pt}
\centering
\scalebox{1.0}{
    \begin{tabular}{*{5}{c}}
      \toprule[0.5mm]
      Method & MAD & MSE & Grad & dtSSD \\
      \midrule
      MODNet\cite{modnet} & 9.50 & 4.33 & 16.94 & 6.22 \\
      RVM\cite{rvm} & 15.42 & 9.22 & 18.34 & 6.95 \\
      VMFormer\cite{vmformer} & 18.39 & 13.12 & 21.57 & 9.10 \\
      AdaM\cite{adam} & 5.94 & 2.79 & 16.71 & 5.45 \\
      MaGGIe\cite{maggie} & 5.78 & 2.66 & 16.18 & 5.24 \\
      \midrule
      OAVM (Ours) & \textbf{5.44} & \textbf{2.48} & \textbf{14.89} & \textbf{5.26} \\
      \bottomrule[0.5mm]
    \end{tabular}}
\caption{\label{crgnncompare}The quantitative evaluation on CRGNN benchmark for evaluating the real-world generalization.}
\end{table}

\begin{table}[t]
\setlength{\tabcolsep}{5pt}
\centering
    \scalebox{1.0}{
    \begin{tabular}{*{7}{c}}
      \toprule[0.5mm]
      \multicolumn{3}{c}{Components} & \multirow{2}{*}{MAD} & \multirow{2}{*}{MSE} & \multirow{2}{*}{Grad} & \multirow{2}{*}{dtSSD} \\
      \cmidrule(lr){1-3}
      OQG & OGCR & SFM \\
      \midrule
      - & - & - & 5.36 & 0.58 & 5.21 & 1.41 \\
      \checkmark & - & - & 5.27 & 1.30 & 5.21 & 1.55 \\
      \checkmark & \checkmark & - & 4.37 & 0.37 & 4.97 & 1.33 \\
      \checkmark & \checkmark & \checkmark & \textbf{4.23} & \textbf{0.31} & \textbf{3.87} & \textbf{1.31} \\
      \bottomrule[0.5mm]
    \end{tabular}}
\caption{\label{ab_net}Ablation study for core components of our method. OQG: Object-level Query Generation. OGCR: Object-Guided Correction and Refinement module. SFM: Sequential Foreground Merging augmentation.}
\end{table}

We further utilize video sequences from DAVIS \cite{davis} and MOSE \cite{MOSE} for comparative studies. These datasets, which were not specifically created for video matting, feature complex scenes that present significant challenges. The qualitative results in Figure \ref{real_visual} demonstrate the superiority of OAVM, showing great generalization and robustness. More experimental results, including initial masks and inference speed, are shown in the supplementary.
\subsection{Ablation Study and Analysis}

\begin{figure}[t]
\centering
\includegraphics[width=0.98\columnwidth]{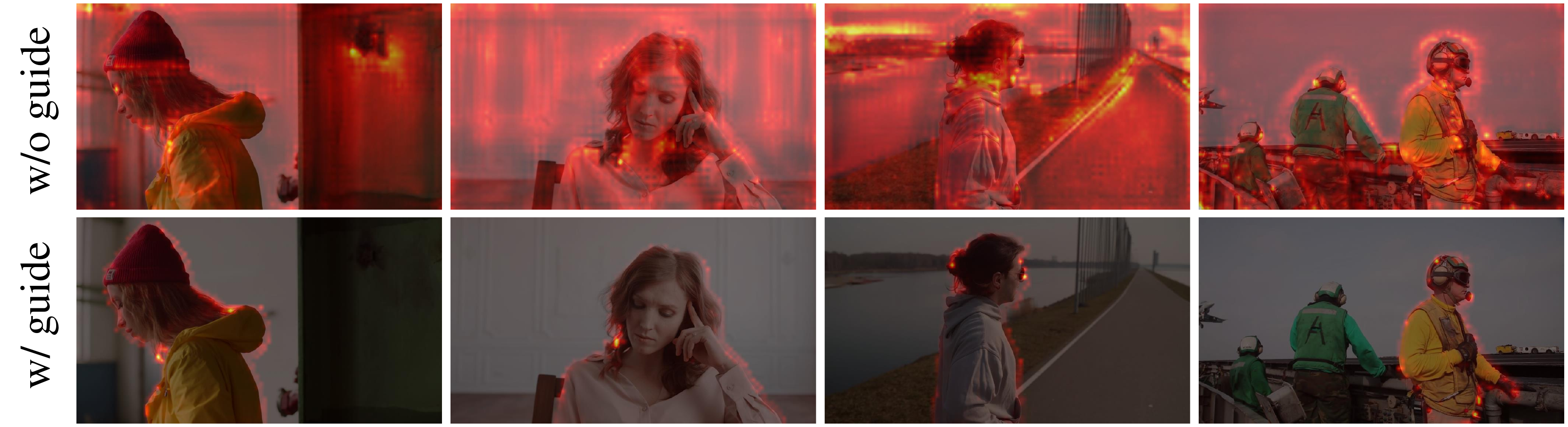}
\caption{Visualization of attention map in the OGCR module. Top: without the cross-frame guidance. Bottom: with the cross-frame guidance. The model can accurately localize foreground objects and handle multi-object scenes with cross-frame guidance}
\label{attn_map}
\end{figure}

\begin{figure}[t]
\centering
\includegraphics[width=0.96\columnwidth]{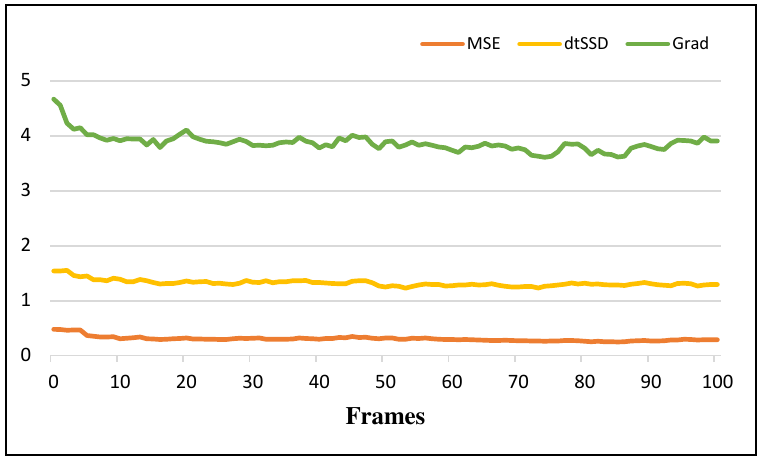}
\caption{Average performance of OAVM over time. This demonstrates the high accuracy and stable temporal coherence of our method.}
\label{line_chart}
\end{figure}

In this section, we conduct ablation studies on the RVM benchmark to analyze the main components and hyper-parameters of OAVM.

\paragraph{Effectiveness of Main Componenets.}

As shown in Table \ref{ab_net}, simply adding the OQG module to enable the network to distinguish between objects leads to a certain performance degradation due to the lack of a reasonable fusion of object-level and pixel-level information. With the usage of our proposed OGCR module, the different information can synergize to deliver better results. The augmentation strategy also helps in learning to distinguish objects.

\paragraph{Visualization of Attention Map.}

The attention map visualization is also performed to evaluate the role of the OGCR module in Figure \ref{attn_map}. It is observed that the network attends globally to almost all pixels without the cross-frame guidance. On the contrary, the network is able to accurately localize the foreground object and focus on the edge regions to extract fine alpha mattes with the guidance. Moreover, as shown in the last column, OAVM efficiently handles multi-object scenes, demonstrating potential in instance matting tasks, which is further discussed in the supplementary.

\paragraph{Average Performance over Time.}

We calculate the average MSE, dtSSD and Grad of OAVM along time steps and the results are presented in Figure \ref{line_chart}. The performance gradually improves at the start of the video and then remains stable. This demonstrates the high accuracy and stable temporal coherence of our method. To further illustrate the powerful capabilities of the OAVM, we provide a demo video and qualitative comparisons over time in the supplementary.

\paragraph{Kernel Size of Dilation.}

The kernel size of the dilation operation in the cross-frame guidance is modified. The outcomes in Table \ref{ab_kernel} show that the effect of kernel size is relatively modest. Note that this experiment was conducted using the first stage for training. Performance drops when the kernel is too large, which we attribute to the fact that too large of the mask region affects the judgment of the target object by including a large amount of background in the foreground.

\begin{table}[t]
\setlength{\tabcolsep}{10pt}
\centering
    \scalebox{1.0}{
    \begin{tabular}{*{5}{c}}
      \toprule[0.5mm]
      \emph{ks} & MAD & MSE & Grad & dtSSD \\
      \midrule
      3 & \textbf{5.07} & \textbf{0.64} & \textbf{0.56} & 1.30 \\
      \midrule
      5 & 5.10 & 0.67 & 0.58 & \textbf{1.29} \\
      7 & 5.24 & 0.76 & 0.64 & 1.32 \\
      \bottomrule[0.5mm]
    \end{tabular}
    }
\caption{\label{ab_kernel}Ablation study of dilation kernel size \emph{ks}.}
\end{table}

\begin{table}[t]
\setlength{\tabcolsep}{8pt}
\centering
    \scalebox{1.0}{
    \begin{tabular}{*{5}{c}}
      \toprule[0.5mm]
      Backbone & MAD & MSE & Grad & dtSSD \\
      \midrule
      MobileNet & 5.07 & 0.64 & 0.56 & 1.30 \\
      \midrule
      ResNet & 4.94 & 0.56 & 0.48 & 1.18 \\
      \bottomrule[0.5mm]
    \end{tabular}
    }
\caption{\label{ab_backbone}Ablation study of backbone.}
\end{table}

\paragraph{Backbone.}

Our OAVM can extract richer sequence features after the backbone is switched to ResNet50 \cite{resnet} as expected, resulting in improved performance. Note that this experiment was conducted using the first stage for training. The results are listed in Table \ref{ab_backbone}.
\section{Conclusion}
This paper explores empowering video matting networks with the ability to understand object-level information explicitly and proposes the OAVM framework. The Object-Guided Correction and Refinement module is proposed, which is able to utilize cross-frame guidance to promote foreground extraction. The Sequential Foreground Merging augmentation strategy is also designed to enrich the sequence scenarios and improve the training pipeline. Extensive experimental results demonstrate that our approach achieves the state-of-the-art performance.


{
    \small
    \bibliographystyle{ieeenat_fullname}
    \bibliography{main}
}

\end{document}